\documentclass[letterpaper, 10 pt, conference]{IEEEtran}  

\IEEEoverridecommandlockouts                              

\usepackage{graphicx}
\usepackage{csquotes}
\usepackage{multirow}
\usepackage{cite}
\usepackage{lipsum}

\title{
Towards Game-based Metrics for Computational Co-Creativity
}

\author{
\IEEEauthorblockN{Rodrigo Canaan}
\IEEEauthorblockA{\textit{Tandon School of Engineering} \\
\textit{New York University}\\
Brooklyn, NY, USA \\
rodrigo.canaan@nyu.edu}
\and
\IEEEauthorblockN{Stefan Menzel}
\IEEEauthorblockA{
\textit{Honda Research Institute}\\ \textit{Europe GmbH}\\ 
Offenbach/Main, Germany \\
stefan.menzel@honda-ri.de}
\and
\IEEEauthorblockN{Julian Togelius}
\IEEEauthorblockA{\textit{Tandon School of Engineering} \\
\textit{New York University}\\
Brooklyn, NY, USA \\
julian.togelius@nyu.edu}
\and
\IEEEauthorblockN{Andy Nealen}
\IEEEauthorblockA{\textit{Tandon School of Engineering} \\
\textit{New York University}\\
Brooklyn, NY, USA \\
nealen@nyu.edu}
}

\begin{document}

\maketitle
\thispagestyle{empty}
\pagestyle{empty}



\begin{abstract}

We propose the following question: what game-like interactive system would provide a good environment for measuring the impact and success of a co-creative, cooperative agent? Creativity is often formulated in terms of novelty, value, surprise and interestingness. We review how these concepts are measured in current computational intelligence research  and provide a mapping from modern electronic and tabletop games to open research problems in mixed-initiative systems and computational co-creativity. We propose application scenarios for future research, and a number of metrics under which the performance of cooperative agents in these environments will be evaluated.\\



\end{abstract}


\section{INTRODUCTION}


Designing intelligent agents characterized by a co-creative, cooperative behavior would mark a major breakthrough in the age of industrial man-machine interaction. Exchanging relevant information with suitable time frequency and enriching the partner (human or machine) with novel perspectives and solution strategies on the problem are key factors for desirable results (considering the value of the output and the effort required). Cooperative games offer the valuable opportunity to realize an interactive environment for developing and evaluating computational methods used by these agents. 

In this paper we review concepts and implementations of cooperative games in the light of their capability to impact development processes in (industrial) environments with co-evolution and co-creativity as important expressions for cooperation. Having a working definition of computational creativity, and how creative systems and their outputs are judged in terms of their value, novelty, interestingness, and surprise, will help us understand cooperatively creative agents and might help us build them as well.  Computational creativity and AI-assisted design are important application areas for computational intelligence techniques such as neural networks, reinforcement learning and evolutionary computation; further, the conceptualization of creativity as search in a design space fits well with design applications of evolutionary computation.

Essentially, this paper tries to answer the following question: what game-like interactive system would provide a good environment for measuring the impact--and success-- of a co-creative, fully cooperative agent?

We begin with a survey of the definition of computational creativity-related terms in the literature, how they relate to each other and how they apply to future work on our own co-creative agents in Section~\ref{sec:creativity}. When considering cooperation between multiple actors (be they human or machine), in addition to the abilities and characteristics of each individual, the attributes of the relationships between individuals and the surrounding environment also impact the success of the endeavor. Section~\ref{sec:mixed-initiative} explores some of these relational or environmental attributes of creative efforts, such as the exchange of information and the share of responsibility. In section~\ref{sec:scenarios} we propose a set of metrics under which to evaluate cooperative agents in game-like environment, and section~\ref{sec:forward} gives our vision of how cooperative agents integrating all discussed techniques should operate in the long term.

\section{COMPUTATIONAL CREATIVITY}
\label{sec:creativity}

Creativity is often understood as the production of novel and valuable concepts~\cite{boden1998creativity}. Computational creativity is a subfield of Artificial Intelligence (AI) that focuses on computational systems whose behavior can be deemed creative~\cite{colton2012computational}. While much theoretical and practical work exists on systems that aim to be creative in their own right, with little or no human intervention~\cite{mccorduck1991aaron, colton2008creativity, colton2012painting, guckelsberger2017addressing}, there are also many systems designed to cooperate with humans to achieve better results than either can presently do alone~\cite{yannakakis2014mixed, kruger2017tools, burstein1996issues}. We focus on concepts of computational creativity and how they map to game-based tasks to further propose a number of concrete game-based metrics for co-creativity in a computational setting.


\subsection{Novelty, Interestingness, Surprise}

In his CSF framework~\cite{wiggins2006preliminary}, Wiggins says an artifact produced by a system is novel if there are no previously existing similar or identical artifacts in the context in which the artifact is produced. Ritchie~\cite{ritchie2007some} builds upon Wiggins' work  and introduces the notion of the inspiring set as the \enquote{knowledge base of known examples which drives the computation within the program}. Ritche calls an artifact generated by the program novel if it is not part of the inspiring set (or not too similar to its members). Both authors admit the possibilities of Novelty being either an absolute assessment (based on the existence of identical artifacts) or, more flexibly, to depend on some metric that establishes degrees of similarity between objects. 

Reehuis \textit{et al.}~\cite{reehuis2013novelty} provide an overview of Novelty metrics used by researchers, and propose dividing them between distance-based metrics and learning-based metrics. Distance-based metrics depend only on the distance, in a specified metric space, between a candidate solution and the archive of earlier solutions (what Ritchie would call the inspiring set). They define uniqueness as the minimum distance between a solution and a member of the archive, as used by~\cite{saunders2010curious} and~\cite{hester2012intrinsically}. Sparseness is defined as the average distance from a candidate solution to its k nearest neighbors in the archive, as used by Lehman and Stanley~\cite{lehman2011abandoning}. Reehuis \textit{et al.} note that uniqueness is equivalent to sparseness with a value of k = 1.

Learning-based metrics take the agent’s expectations into account. Formally, let an agent (or an external viewer) be imbued with a model of the world, which  ascribes probabilities to certain events. High novelty, or surprise occurs when the agent comes into contact with examples which contradict the model. Reehuis \textit{et al.} provide the prediction error, dispersion in predictions and predictive variance of the model as examples of learning-based novelty. Itti and Baldi~\cite{itti2009bayesian} provide a bayesian deﬁnition of surprise using the relative entropy, or Kullback-Leibler (KL) divergence~\cite{kullback1997information}. Since the KL divergence depends on a prior probability distribution, we could also classify it as learning-based novelty.

\subsection{Analysis of distance- and learning-based novelty metrics}

\begin{figure}
\centering
\includegraphics[width=0.5\textwidth]{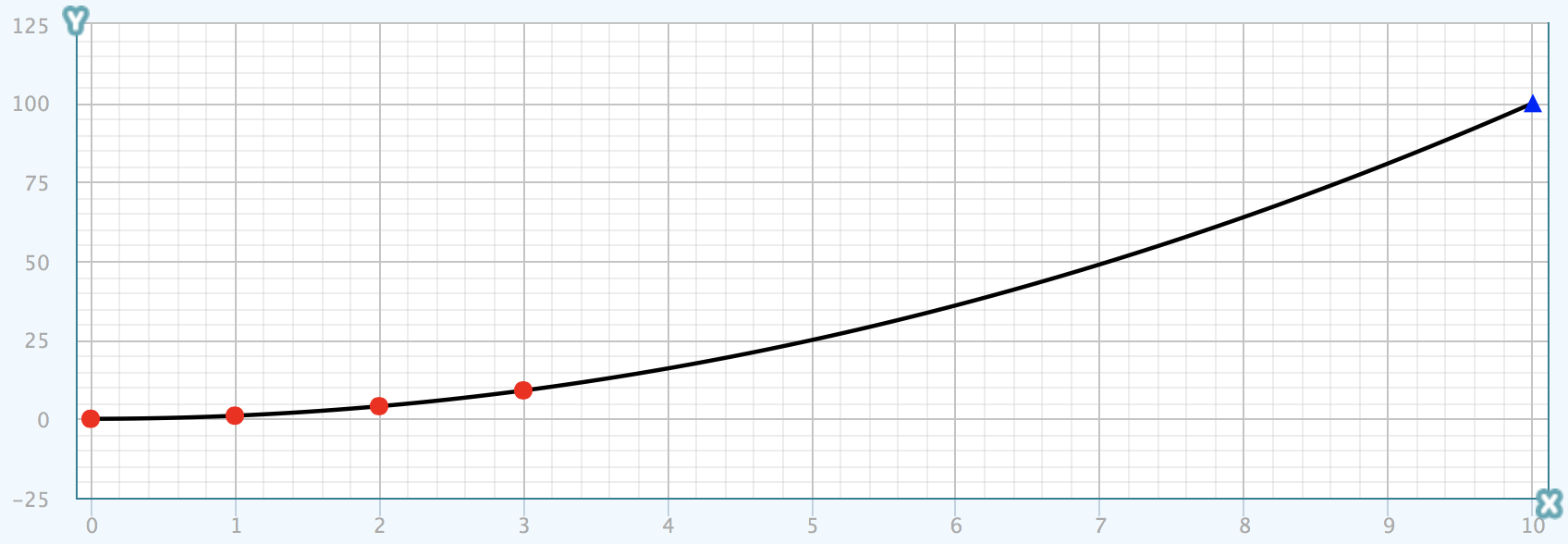}
\caption{\label{fig:Parabola_Fit}A simple polynomial regression model trained on the dataset $I = [(0,0),(1,1),(2,4),(3,9)]$ (in red) would perfectly predict the point $x = (10,100)$  in blue, even though the Euclidean distance to the inspiring set is large}
\end{figure}

We provide a simple example of the distinction between the two kinds of novelty in figure \ref{fig:Parabola_Fit}. The points in red are part of the inspiring set $I = [(0,0),(1,1),(2,4),(3,9)]$ and a candidate solution $x = (10,100)$ is shown in blue. A naive Euclidean distance-based metric would ascribe high novelty to x, while a simple learning model based on polynomial regression could might ascribe zero novelty to x, since it is a perfect fit to the parabola $y=x^2$. Thus, under learning-based novelty, what is novel to one observer might not be to another.

It is clear to us that the distinction between distance and learning-based novelty is didactic only. A high novelty value in a distance-based metric such as sparseness or uniqueness is equivalent to a low probability in a simple model that takes only the Euclidean distance from the points in the inspiring set into account (with more distant points being less probable). On the other hand, a more complex learning model can be abstracted as a distance metric in a sufficiently high dimension.

Thus, the choice of novelty metric to use depends on the problem. If one must describe a model being refined over time, or multiple agents with individual models making different predictions, a learning-based metric might be ideal. If there is no explicit model, or a single static model and  a distance metric is readily available, it might be preferred. Richter~\cite{richter2016analyzing} defines a \enquote{neighborhood structure} as an integral part of a fitness landscape, so we believe evolutionary computation is a good environment for distance-based metrics.

Whatever the kind of metric used, is important to note that a higher value of novelty is not necessarily desirable. As a simple example, consider a set of observations consisting entirely of random noise, such as a \enquote{poem-generator} that simply generates long strings of random characters. These would have high novelty (either in the distance or learning sense), but it could hardly be called a poem generator. It is clear that both  low-novelty and extreme-novelty can be undesirable to a system, which is why some authors define the \textit{interestingness} of an object as a function relating its novelty to some desirability metric. A Wundt curve \cite{wundt1874grundzuge} is a hedonistic function commonly used to express this relationship~\cite{saunders2010curious} \cite{reehuis2013novelty} \cite{graziano2011artificial}. In this sense, interestingness might be characterized by just the right amount of novelty - not too much, not too little.

\begin{figure}
\centering
\includegraphics[width=0.39\textwidth]{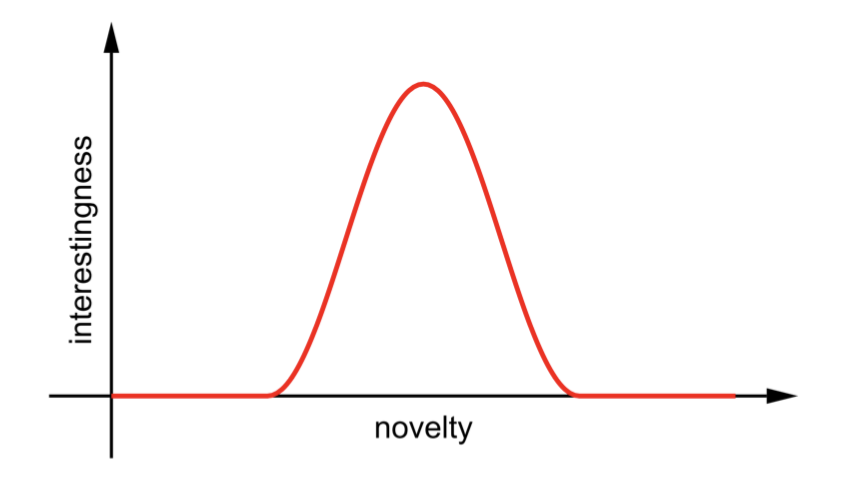}
\caption{\label{fig:Wundt_Curve}A Wundt Curve, as shown in~\cite{saunders2010curious}}
\end{figure}

Learning-based interestingness is also defined in a way to avoid excessively high novelty (unpredictability). Schmidhuber, as part of his theory of artificial curiosity, provides a comprehensive framework~\cite{schmidhuber2010formal} for characterizing the learning progress of an agent by noting the intimate relationship between prediction and compression. An observation is termed interesting if it enables the agent to learn some previously unknown irregularity, that is, further compress the available data. Rehuis \textit{et al.}~\cite{reehuis2013novelty} discuss a number of different learning-based interestingness metrics, which attempt to maximize the learning progress induced by including a new observation in the model: Actual Learning Progress, Previous Learning Progress, Previous Competence Change and Reducible Error. These are all based on the difference between the prediction error in a region of the problem space at two points in time.

The use of these terms (novelty, interestingness, value) is not entirely consistent across all literature.
For this reason, we find it convenient to settle on some definitions for our purposes, which lean closer to the way the terms are used in \cite{reehuis2013novelty}. These definitions are:

\textbf{Novelty:} any measure of dissimilarity between a sample concept and a collection of concepts (distance-based novelty) or an expression of the prediction error of a surrogate model (learning-based novelty).

\textbf{Interestingness:} a function of how desirable a solution is based on its novelty. This will typically assign a low score both to low-novelty and excessive-novelty solutions. 

\textbf{Surprise:} synonymous with learning-based novelty, that is, a measure of how much a candidate solution deviates from a model's expectation.

\subsection{Value}

Wiggins defines Value as \enquote{The property of an artifact (abstract or concrete) output by a creative system which renders it desirable in the context in which it is produced}. Given that we also defined interestingness with regards to desirability, a closer look at the relationship between interestingness and value is necessary.

We define value as any measure of desirability, possibly domain-specific, while interestingness will be used solely as a more domain-agnostic measure of desirability that depends only on the underlying novelty metric and possibly the agent's internal state, but not on any externally assigned goals. To make the distinction clear, we propose an example inspired the space probe described in \cite{graziano2011artificial}. Imagine a space probe designed for mining some kind of ore in a distant planet. It has a number of sensors to measure some features of the world and is able of movement in four different directions. Via reinforcement learning, it uses data from its sensors to build a model that predicts the concentration of ore in parts of the world.

Consider now two regions of the world $R1$ and $R2$. At some point in time the model predicts a high concentration of ore in $R1$ and low concentration in $R2$. After exploring both regions, $R1$ is found to have low concentration of ore, $R2$ is found to have a high concentration and the model is updated. From a pure learning perspective (that is, in terms  of learning progress), both observations can be equally useful. From a value perspective, it is clear that $R2$ has more value. $R1$ is only useful to the extent that, by exploring similar regions, the probe might eventually learn a new pattern that enables it to avoid such low-value regions in the future.

As Graziano puts it, a reinforcement learning agent can be given an \enquote{internal or curiosity reward}, which directs its learning, and an \enquote{external reward}, defined to achieve some pre-defined goal.
These must be balanced against each other, as, unless the agent is provided with an accurate model from the start, it first needs to learn where the high-value regions are by exploring unknown (possibly low-value) regions. This is known as the exploration and exploitation problem. 

A more classic formulation of the exploration and exploitation problem is given by the Multi-Armed Bandit (MAB) problem, in which a gambler is faced with $N$ slot machines (also known as \enquote{one-armed bandits}) with unknown reward distributions and must decide which machine to play at each point in time. An in-depth study of the MAB problem is outside the scope of this article. For more information, see~\cite{agrawal2012analysis}. In a Reinforcement Learning context, we will take  novelty or interestingness (depending on the formulation of the problem) to be related to an agent's internal reward, encouraging exploration, and value to be related to an agent's external objective, encouraging exploitation. For a pure learning agent, an external definition of value might not be necessary.

Another interesting application of the relationship between novelty and value is seen in Lehman and Stanley's novelty-based evolution~\cite{lehman2011abandoning}. They implement novelty search as an extension of the NEAT method~\cite{stanley2002evolving}, using sparseness  as metric for novelty, where distance is a domain-dependent measure of behavioral difference. Sparseness is, in turn, measured against the current population plus an archived set of high-novelty solutions. The novelty of a solution is used as selection factor for the evolving population, and the external objective is only used as a stopping condition test. By not using a fitness function based on the external objective, they outperform traditional methods in some deceptive environments, that is, where the fitness function leads too often to local optima. This indicates that when a good heuristic for the desired objective  is unavailable, search through novelty alone can still lead to good results. Another possibility is a combined approach, where both novelty and traditional fitness are rewarded concurrently in a multi-objective formulation of the problem~\cite{mouret2011novelty}.

\section{Games as Mixed-Initiative Research Platforms}
\label{sec:mixed-initiative}


The recent years have seen advancements both in systems that facilitate human creation and systems able of autonomous creation. However, researchers have noted a gap in systems that can work in tandem with one or more human agents, and achieve similar levels of initiative and responsibility as would be expected from a human partner. These are known as mixed-initiative systems. Some authors also use the term  human-computer co-creativity, or mixed-initiative co-creativity, when emphasizing the creative nature of the output of such systems.


Carbonel~\cite{carbonell1970mixed} defines mixed-initiative systems as "one in which both humans and machines can make contributions to a problem solution, often without being asked explicitly". This notion is developed by Burstein and Mcdermott~\cite{burstein1996issues}, who investigate how humans and machines can "best share information about and control of plan development" in a mixed-initiative system so that each agent works in areas where they perform best, use the appropriate representation for the communication of plans and have the means of acquiring and transferring authority over tasks. They identify six areas of AI research that needed to be addressed to enable their proposed model of mixed-initiative planning systems: plan-space search control management, representations and sharing of plans, plan revision management, planning and reasoning under uncertainty, learning and inter-agent communications and coordination.

Yannakakis \textit{et al}.~\cite{yannakakis2014mixed} identify \enquote{mixed-initiative co-creation (MI-CC) as the task of creating artifacts via the interaction of a human initiative and a computational initiative}, emphasizing the proactivity of the contributors, and differentiating it from \enquote{non-proactive computer support tools (e.g. spell-checkers or image editors)}. They also argue that, if such a system is able to foster human creativity, then it can be called mixed-initiative co-creativity.

Kr\"{u}ger \textit{et al.}~\cite{kruger2017tools}  classify interaction between human and machine system in three levels of cooperation complexity: tools, adaptive tools and cooperative assistants. With a \enquote{tool} the human user has complete responsibility for the success of the operation and adaptation to different tasks. An \enquote{adaptive tool} has a model the environment to adapt to different situations, but has no capability to resolve possible mismatch between its goals and the humans goals. \enquote{Cooperative assistants} have a model not only of the environment, but of the human user, and are equipped with a transparent interface enabling the negotiation of responsibilities and goals. Although they do not use the term mixed-initiative, it is our view that such a cooperative assistant would qualify as mixed-initiative.

A similar distinction is drawn by Davis \textit{et al.}~\cite{davis2014building}, between what they call Creativity Support Tools (which support a creative person), Computational Creativity systems (which autonomously create products) and Computer Colleagues, which are \enquote{Co-creative agents (that) collaborate with humans in real-time improvisation to enrich the creative process}. Davis~\cite{davis2013human} previously defined human-computer co-creativity as \enquote{a situation in which the human and computer improvise in real time to generate a creative product}, where \enquote{the contributions of human and computer are mutually influential} and that \enquote{introduces a computer into this collaborative environment as an equal in the creative process}. (Though one can of course think of useful co-creative processes where the computer is not an equal.)

Games have been considered the \enquote{killer app} for computational creativity~~\cite{liapis2014computational},  due to being multifaceted , content intensive, benefiting greatly from procedual generation techniques and rich (highly interactive and engaging). Games have also traditionally been used as benchmarks for AI. Of particular interest are general game-playing algorithms, which can in principle be applied to any games and better generalize to other real-world problems. For example, the GVGAI competition offers a set of 2D arcade-like games~\cite{perez20162014}. The use of games as AI benchmarks has received recent media attention due to the success of DeepMind's success at the game of Go with AlphaGo~\cite{AlphaGo}, AlphaGo Zero~\cite{AlphaGoZero} by combining reinforcement learning and Monte Carlo tree search. This paradigm has also yielded success in other games by Anthony \textit{et al.}~\cite{anthony2017thinking} and by DeepMind's AlphaZero~\cite{AlphaZero}. 
Games are also fun. Perez \textit{et al.}~\cite{perez20162014} suggest that this leads to higher interest in AI research by the general public, and a 2014 review of gamification studies by Hamari \textit{et al.}~\cite{hamari2014does} concludes that, although some methodological issues were found, most studies yielded positive effects of gamification. We would like to investigate whether the use of game-like techniques can lead to the design of better co-creativity tools for real world problems.

Finally, we have identified several modern games where we believe a good AI controller, especially one designed for co-operative play with humans, would benefit from addressing many specific issues listed by mixed-initiative and co-creativity researchers as research topics for the development of the field. Tables~\ref{Research_topics_games1} and ~\ref{Research_topics_games2} illustrate a mapping between these research topics and games that would serve as interesting problems for those research topics. We further detail the correspondence between research topics and games below:

\begin{table*}[]
\centering
\caption{Research topics and games}
\label{Research_topics_games1}
\begin{tabular}{|c|c|c|l|l|}
\hline
\multirow{2}{*}{\textbf{Agent Modelling}}                                                                       & \multirow{2}{*}{\textbf{Changing Environment}}                                                                         & \multicolumn{3}{c|}{\textbf{Nontrivial Goals}}                                                                                                                                                                                                                                                                      \\ \cline{3-5} 
                                                                                                                &                                                                                                                        & \textbf{Emerging goals}                                                                    & \multicolumn{1}{c|}{\textbf{Hidden Goals}}                                                                        & \multicolumn{1}{c|}{\textbf{Dynamic Goals}}                                                        \\ \hline
\multicolumn{1}{|l|}{\begin{tabular}[c]{@{}l@{}}Poker\\ Race for the Galaxy\\ \textit{Hanabi}\\ \textit{Magic Maze}\end{tabular}} & \multicolumn{1}{l|}{\begin{tabular}[c]{@{}l@{}}\textit{Pandemic}\\ \textit{Flashpoint}\\ \underline{\textit{Overcooked}}\\ \end{tabular}} & \multicolumn{1}{l|}{\begin{tabular}[c]{@{}l@{}}\underline{Minecraft}\\ Roleplaying Games\end{tabular}} & \begin{tabular}[c]{@{}l@{}}Mafia\\ Werewolf\\ The Resistance\\ \textit{Shadows over Camelot*}\\ \textit{Dead of Winter*}\end{tabular} & \begin{tabular}[c]{@{}l@{}}Ticket to Ride\\ Terra Mystica\\ \textit{Pandemic Legacy: Season 1}\end{tabular} \\ \hline
\end{tabular}\\
\centering{Mapping of research topic to games. Games in italics are cooperative. Games with an asterisk are cooperative with an optional traitor mechanic. \\ Underlined games are electronic games}
\end{table*}

\begin{table*}[]
\centering
\caption{Research topics and games (cont.)}
\label{Research_topics_games2}
\begin{tabular}{|c|c|l|}
\hline
\multirow{2}{*}{\textbf{Asymmetric responsibilities}}                                                   & \multicolumn{2}{c|}{\textbf{Communication}}                                                                                       \\ \cline{2-3} 
                                                                                                       & \textbf{Unconstrained}        & \multicolumn{1}{c|}{\textbf{Constrained}}                                                         \\ \hline
\multicolumn{1}{|l|}{\begin{tabular}[c]{@{}l@{}}\textit{Pandemic}\\ \textit{Magic Maze}\\ \underline{\textit{Can't Drive This}}\end{tabular}} & \multicolumn{1}{l|}{\begin{tabular}[c]{@{}l@{}}\textit{Pandemic}\\ \textit{Flashpoint}\end{tabular}} & \begin{tabular}[c]{@{}l@{}}\textit{Hanabi}\\\textit{Magic Maze}\\ Real-Time Games (in general)\\ Competitive Games (in general)\end{tabular} \\ \hline
\end{tabular}
\end{table*}


\textbf{Agent modeling:}
A lot of research in mixed-initiative systems and co-creativity is concerned with building a good model of the other agent's behavior and goals. For Burstein and Mcdermott~\cite{burstein1996issues}, intent recognition (e.g. filling in the gaps of a plan that is not specified to the degree of atomic actions) and learning user preferences are important tasks of mixed-initiative planing systems. The ability to build a model of the user is one of the factors that distinguish a cooperative assistant from an adaptive tool for Kr\"{u}ger \textit{et al.}~\cite{kruger2017tools}. 

Hadfield-Menel \textit{et al.}~\cite{hadfield2016cooperative} introduce Cooperative Inverse Reinforcement Learning (CIRL), a framework of cooperation between a Human $H$ and a robot $R$, where both players are rewarded by the same reward function, which is known only by $H$. $R$ tries to infer the reward function from $H$'s actions. They show that when $H$ tries to greedily maximize its own rewards, $R$ might learn a poor approximation to the real reward function and achieve suboptimal results, so optimal solutions may involve active instruction by the human. The use of Generative Adversarial Networks (GANs)~\cite{goodfellow2014generative} to generate novel artifacts based on the design objectives of a user~\cite{bontrager2018deep} or emulating a specific art style~\cite{elgammal1706can} is also a recent and promising approach to this problem.

The amount of time or data available for learning  can also impose constraints on the techniques used. If a behavior must be learned over the course of a single game session, for example (rather than over a large number of games), one approach used by Barret \textit{et al.}~\cite{barrett2011empirical} is to pre-compute a set of strategies $S$ and assume the other player is using strategy $S_i$ with probability $p_i$, using Bayesian reasoning to update the probability of each strategy whenever the other player makes an action. The value of a prospective action with each possible paired strategy is weighted by their probability to determine the best action. They show this can lead to better results than simply mirroring the other player, even when the actual strategy is not one of the strategies contained in $S$.

Another useful technique is empowerment maximization~\cite{salge2014empowerment, guckelsberger2016supportive}. Empowerment is an information-theoretic, intrinsic motivation metric that formalizes how much potential causal influence an agent has upon the world it can perceive. An artificial agent motivated to maximize its human partner's empowerment could sidestep the issue of creating a complex model of the other agent's intentions by simply acting to leave their partner's options open.

In games, the need to predict the other player's actions and objectives arises naturally in competitive environments, especially those involving simultaneous action selection (like \textit{Race for the Galaxy} (Tomas Lehmann, 2007) and other forms of bluffing (like Poker). In cooperative games, the need for agent modeling is alleviated if players are allowed to freely coordinate their actions. However, some cooperative games like \textit{Hanabi} (Antoine Bauza, 2010) and \textit{Magic Maze} (Kasper Lapp, 2017) enforce communication restrictions, which makes agent agent modeling a key factor for success.

\textbf{Changing environment:}
Referring to traditional AI planning systems, Burstein and Mcdermott~\cite{burstein1996issues} state \enquote{the worlds in which these planners worked tended not to change much, fight back at all}, and regards plan revisions and reasoning under uncertainty as two major areas of necessary research. For Kr\"{u}ger \textit{et al.}~\cite{kruger2017tools}, the ability to\enquote{change one
or more of its own parameters in response to environmental
variations} separates regular tools from adaptive tools, and is one of the requirements for cooperative assistants.

Many modern tabletop games excel in thematically representing environment changes inspired in real world uncertainties. In \textit{Pandemic} (Matt Leacock, 2008), in during the Infection phase, cards are drawn from and infection deck to randomly add disease cube to cities in the board. If not treated timely by the players, these might induce chain reactions and defeat the players.  \textit{Flashpoint: Fire Rescue} (Kevin Lanzing, 2011) has the Advance Fire phase, where smoke and fire can be added to the board, which can cause explosions, structural damage to a collapsing building and knock down player-controlled Firefighter units. These phases usually occur in between player action phases to randomly provide either resources or obstacles to the players, and we term them environment phases for generality. In \textit{Overcooked}, an electronic cooperative cooking game, the ingredients each player has access to changes with shifts in the map layout.

In some games, the goal of the game itself (that is, the scoring function) may change unpredictably during the course of the game (for example, limited-time scoring opportunities). We investigate these and other goal-related features below.

\textbf{Nontrivial goals:}
Real-life goals are often nontrivial. They might be unknown to some of the agents, as in~\cite{hadfield2016cooperative}. The goal might change during the execution of a project or parts of it may be implicitly specified~\cite{burstein1996issues}. The goal might be complex and broken into subtasks, and the responsibility for each subtask must be properly assigned, which could involve negotiation~\cite{burstein1996issues,kruger2017tools}. In short, Davis \textit{et al.}~\cite{davis2014building} characterize goals as \enquote{socially negotiated, dynamic and emergent}.

In some \enquote{games}\footnote{At this point, we want to  acknowledge the controversy in calling these activities games. In \textit{Rules of Play}~\cite{salen2004rules}, Salen and Zimmerman's definition of game involves there being a quantifiable outcome. We sidestep this discussion and call them games for simplicity and consistency with common usage.}, such as \textit{Minecraft} (Mojang, 2008) and roleplaying games, there is no overall objective stated by the rules, although the players might still define objectives for themselves based on what is fun for them, negotiate it with other players and attempt to achieve them via cooperation or competition. We term these games with regards to their goal as Emerging, due to their emergent nature as a product of the interaction between players and the environment.

Modern tabletop also employ many variations of secret objectives.  Although we could not find a fully cooperative game with hidden goals, social deduction games such as \textit{Werewolf} (Davidoff, Plotkin, 1986) and \textit{The Resistance} (Don Eskridge, 2009) feature competition between two or more factions (whose members cooperate among themselves), where each player typically only know the allegiance of a small fraction of the other players (and thus, their objectives). \textit{Shadows Over Camelot} (Cathala, Laget, 2005) and \textit{Dead of Winter: A Crossroads Game} (Gilmour, Vega, 2014) are semi-cooperative games with a random probability of one player being assigned a traitor role. The mere possibility of a traitor encourages players to second-guess other player's reasons. Dead of Winter features a fairly unique mechanic where, even if no traitor is present, each player's goal is composed of a public objective, shared by all non-traitor players, and a secret objective, where a player only wins if the group fulfills the public objective and they personally fulfill their secret objective (so that one or more players might still lose even if the group achieves success). This adds another layer of complexity where seemingly strange behavior by a player can be justified either by their secret objective or by a traitor role, and a non-traitor player's need to fulfill their secret objective might lead to the failure of the entire group.

Dynamic goals (where the scoring function itself changes over the course of a game) are also common: in \textit{Ticket to Ride} (Alan Moon, 2004), players have the option of drawing extra objective cards, achieving extra score if they manage to fulfill these new objectives, at the risk of score penalties if they fail. In \textit{Terra Mystica} (Dr\"{o}gem\"{u}ller, Ostertag, 2012), a unique scoring tile is randomly drawn for each turn, enabling limited-time scoring opportunities for all players. A cooperative example is \textit{Pandemic Legacy: Season 1} (Daviau, Leacock, 2015), a variation of \textit{Pandemic} where players play missions in a persistent and evolving world, and a mission's objective may be altered mid-course by specific storyline events.

\textbf{Asymmetric responsibilities and areas of expertise:}
For Burstein and Mcdermott~\cite{burstein1996issues}, two of the high-level goal are to enable proper communication between agents with different areas of expertise, and that each agent works in areas where they perform best. Kr\"{u}ger \textit{et al.}~\cite{kruger2017tools} gives an example of a cleaning robot that is able to identify areas where it cannot access (e.g. due to being blocked by an object) and proactively request assistant from the human user (who has a different set of skills and is able to e.g. move the object away). Different responsibilities (such as teaching and learning) can also be a result of asymmetric information, such as in~\cite{hadfield2016cooperative}.

In \textit{Pandemic} and many other games, each player controls a unique character with special abilities, such as performing one specific type of action more efficiently. Some games are more radical in the variability between player powers. In \textit{Magic Maze}, players share control of a group of character pawns, but each player can only move a pawn in one specific cardinal direction. In \textit{Can't Drive This} (Pixel Maniacs, 2016), one player takes the role of a driver while the other dynamically builds the road on which the first player must drive.

\textbf{Communication:}
Researchers highlight the need for a shared representation~\cite{burstein1996issues} or interface~\cite{kruger2017tools} in which communication can happen. Burstein and Mcdermott~\cite{burstein1996issues} also implicitly acknowledge a cost to associated with communication when stating \enquote{it is almost necessarily the case that details will be left out, if the communication is to be succinct enough to make it worth defining the task for another to carry out}. Lu \textit{et al.}~\cite{lu2017cooperative} use a cooperative co-evolutionary approach to demonstrate how the frequency at which communication occurs impacts cooperative performance under different communication costs. Finally, the problem definition itself might disallow certain forms of communication, or allow no communication at all, in which case agents still can gain information by reasoning about other agents' actions~\cite{hadfield2016cooperative}.

Games offer an avenue for exploring all of these problems. In games that allow unrestricted communication, such as \textit{Pandemic} and \textit{Flash Point}, complex communication involving conditional logic and algorithm building can emerge, as shown by Berland~\cite{berland2012collaborative}. Designing a communication scheme with comparable expressive power for effective human-AI and AI-AI cooperation is an open problem. In the Tiny Coop environment~\cite{waltoncontrolling} communication actions are available, allowing each agent to signal the direction it would like its partner to move in the immediate future. However, human communication often happens not at the level of individual actions, but in terms of higher level goals and their dependencies. The need for communication can also be triggered by specific events, such as the completion of a goal or a change in environment. A recent example of development in this direction are by Schrodt \textit{et al.}~\cite{schrodt2017event}, whose agent is able to establish cooperative goals in a variant of \textit{Super Mario Bros.} while thinking out loud its current intentions and state.

In other games, the communication is restricted by the game rules. In \textit{Hanabi}, players can only communicate by expending a limited number of hints, which can only state the color or value of cards in another player's hand. In \textit{Magic Maze}, players can freely communicate, but only at specific points in time. As a real-time game, time spent elaborating the plan comes at the cost of time for execution of the plan.

In some competitive games, the rules allow full communication, including negotiation and partnerships, but it must occur in full view of other players. In this scenario, the cost of communication is the information that is leaked to antagonist players, and so communication is a strategic decision.

\section{Metrics for Co-creative Agents}
\label{sec:scenarios}

We propose the following types of metrics for co-creative agents in game environments:

\begin{itemize}

\item \textbf{Value: } For any game with fixed objectives stated by the rules, a natural way to measure value is the game's scoring function. For games with emerging or hidden goals, explicit feedback from  the user, if available, can also be used as a value metric. For procedurally generated content, value could be measured by a pre-determined fitness function of the generated artifact's features (as seen in ~\cite{yannakakis2014mixed}), by results of simulation~\cite{isaksen2016statistical} or by subjective evaluation of the human player, who selects their preferred generated artifact~\cite{yannakakis2014mixed}.

\item \textbf{Learning-based metrics: } An agent might attempt to build a model of the other agents over the course of a game session or multiple sessions. A model of the user's behavior can be used to predict their action in a tree search algorithm. A model of the user's preferences can be used to predict the probability of acceptance of an artifact by the user. The accuracy of these predictions is a metric of learning-based novelty, and the higher the confidence of the model in a result, the higher the surprise if the prediction fails. 

For an agent attempting to gradually build a model of a player, care must be taken to isolate gains in performance (either in terms of value or in terms of accuracy of predictions) due to improvements on the part of the agent and improvements on the part of the human. After all, a human player could play with a simple, non-learning agent, and the agent could still report an increase in performance due to the human player better learning to play the game or play matching the agent's expectations. A statistical analysis of player improvement over time is given in~\cite{isaksen2016statistical}.

To avoid this confounding factor, it is important implement baseline agents, with statistically unchanging behavior, who would serve as proper control groups when paired with learning agents and humans, so that the impact of an agent's learning on performance cannot be overestimated.

\item \textbf{Distance-based metrics: } In some scenarios, the product of each decision by the agent will not be a single, atomic action, but a number of options for the other agents to choose from, such as a number of action plans or a number of in-game artifacts for use of the human player. In these cases, distance-based metrics of novelty and interestingness can be used to make sure the suggestions offer a varied sample of the decision space, rather than small variations of a single idea. That way, the user is most likely to find a suggestion they identify it, and tweak some finer details to their own preference.

\item \textbf{Empowerment metrics: } Empowerment "grows when different actions lead to different perceivable outcomes", and is a form of intrinsic motivation \cite{guckelsberger2016supportive}. As such, it can be used in the absence of explicitly stated goals. We believe empowerment can also be used to maximize chance of acceptance of a suggestion, similar to distance-based metrics, by providing many relevant choices to the user.

\item \textbf{Communication metrics: } The most direct way to measure the effectiveness of a communication scheme is simply to measure the difference in value (or in accuracy of predictions) achieved by cooperating under different schemes (or with no communication), as is done in~\cite{waltoncontrolling}. The frequency of communication~\cite{lu2017cooperative} can also be an important metric in scenarios where there is a communication cost or where player experience could be negatively impacted by a high-frequency stream of low-level communication actions.

\end{itemize}

The proposed metrics pose an initial approach to quantify the success of co-creative agents in cooperative games and similar environments.

\section{The way forward}
\label{sec:forward}

In tables~\ref{Research_topics_games1} and ~\ref{Research_topics_games2}, we listed some characteristics of games that provide interesting research topics for human-computer cooperation. In our view, the most promising application scenarios are those focused on \textbf{agent modeling} and \textbf{communications} and are the biggest gap in current cooperative systems. They are at the core of co-creative cooperative activities, while the remaining entries of the table serve	 as challenges to be addressed by better cooperative systems (which include agent modeling and communication as core components): how will the other player react by a change in the environment? How to infer an unknown goal from a player's actions? How to communicate a change of plans due to a change in the environments? How to communicate (or predict) which activities are to be performed by each agent, especially under time constraints?

Going forward, we believe communication and agent modeling can also feed off each other. On one hand, an agent can use its ability to communicate to build more accurate models, either by directly asking for missing information or by picking up on cues from information provided by its peers. On the other hand, having an accurate model can help determine what information to share or ask for. A very clear example of this dynamic is in the game \textit{Hanabi}, where different players are comfortable operating under different levels of implicit information (e.g. how willing are they to risk playing a card with incomplete information?). Observing the hints given by a player can help us infer how much information they need for their own actions, while knowing how they act under uncertainty helps us determine what hints to give.

While section~\ref{sec:mixed-initiative} provides many examples of application scenarios to achieve progress in human-machine cooperation in the short term, our long term view is that this research  can lead to applications where high-level goals and plans can be negotiated between human and artificial agents, taking into account their specific abilities and knowledge. The artificial agents will then be able to fill in small gaps in the plan by reasoning about a model of the world and of the other agents. Alternatively, the agent can proactively request any information it is missing if the gaps are too large to be filled. 

It will be able to detect events that require a change of plans (such as a change in environment, available resources or goals of the group) and once again communicate and negotiate the new plan. All along the process, novel and valuable artifacts will be produced through computational creativity techniques, where novelty and value are judged in regards to a model of the knowledge and preferences of the target audience.

\section*{CONCLUSION}

We started this paper by asking what game-like environments would be ideal for measuring the impact and success of co-creative cooperative agents. We answer that question by proposing several types of metrics, based on a thorough research on computational creativity and metrics used in the computational intelligence community for the related concepts of novelty, value, interestingness and surprise.  We have shown how research in these scenarios, and similar games, can help shed light on open questions of the field and provided a vision of how these systems could operate in the long term.

We hope that this can lead to the development of better mixed-initiative, co-creative systems for a variety of domains, including industrial applications, where human and machine can cooperate working in areas where they perform best, communicating efficiently to achieve nontrivial goals under a changing, uncertain environment.


\section*{ACKNOWLEDGMENT}


Rodrigo Canaan gratefully acknowledges the financial support from Honda Research Institute Europe (HRI-EU). We would also thank Nikolas Dahn (TU Ilmenau) and Dr. Thomas Weisswange (HRI-EU) for their valuable input to this research.

\bibliographystyle{IEEEtran}
\bibliography{IEEEabrv,mybibfile}

\end{document}